# LiteMuL: A Lightweight On-Device Sequence Tagger using Multi-task Learning

Sonal Kumari, Vibhav Agarwal, Bharath Challa, Kranti Chalamalasetti, Sourav Ghosh, Harshavardhana, Barath Raj Kandur Raja

Samsung R&D Institute Bangalore, Karnataka, India 560037

{sonal.kumari, vibhav.a, bharath.c, kranti.ch, sourav.ghosh, harsha.vp, barathraj.kr}@samsung.com

*Abstract—* **Named entity detection and Parts-of-speech tagging are the key tasks for many NLP applications. Although the current state of the art methods achieved near perfection for long, formal, structured text there are hindrances in deploying these models on memory-constrained devices such as mobile phones. Furthermore, the performance of these models is degraded when they encounter short, informal, and casual conversations. To overcome these difficulties, we present LiteMuL – a lightweight on-device sequence tagger that can efficiently process the user conversations using a Multi-Task Learning (MTL) approach. To the best of our knowledge, the proposed model is the first on-device MTL neural model for sequence tagging. Our LiteMuL model is about 2.39 MB in size and achieved an accuracy of 0.9433 (for NER), 0.9090 (for POS) on the CoNLL 2003 dataset. The proposed LiteMuL not only outperforms the current state of the art results but also surpasses the results of our proposed on-device task-specific models, with accuracy gains of up to 11% and model-size reduction by 50%-56%. Our model is competitive with other MTL approaches for NER and POS tasks while outshines them with a low memory footprint. We also evaluated our model on custom-curated user conversations and observed impressive results.**

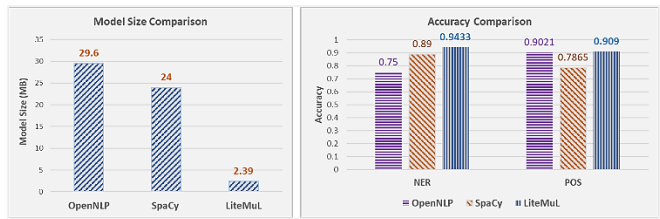

Figure 1. Comparison of our proposed LiteMuL against existing NER and POS models: OpenNLP (On-device, trained on Reuters Corpus), SpaCy (Offline, trained on OntoNotes 5 Corpus), LiteMuL (On-device, trained on CoNLL 2003 and Custom dataset).

## I. INTRODUCTION

Named entity recognition (NER) and part-of-speech tagging (POS) are crucial tasks to enable low-level information extraction in natural language processing (NLP) systems. NER is a sequence tagging task that recognizes named entities such as people names, organizations, locations, etc., from the input text, while POS tagging aims at syntactic annotation of the input text. Together, NER and POS form the fundamental blocks for NLP systems such as information extractions, machine translation, and summarization to name a few.

NER & POS tasks have been well-researched problems and achieved perfect scores on evaluation against standard datasets. But, it is very challenging for these models to show similar performance on short text due to the ambiguous nature of the language and the noise introduced by the informal conversations [1]–[4].

Most of these state of the art deep neural networks are huge pre-trained models, often containing hundreds of millions of parameters. Recently transformer-based models like BERT (12 layer – 110M params) [5], MobileBERT (24 layer – 15.1Mparams) [6], DistilBERT (66M params – 200 MB) [7] shows remarkable results on various NLP applications.

However, these architectures are deep and the network computations are in the order of quadratic, making it challenging to deploy directly on the low-resource electronic devices such as mobile phones, tablets, wearables, etc.

In this paper, we propose an on-device Multi-Task Learning (MTL) model based on a neural network (named as LiteMuL) that enables (i) recognizing named entities in casual conversations (ii) with compact model, and (iii) competing accuracies. Furthermore, to NER recognition, we employ POS as an auxiliary secondary task. The network jointly detects NER and POS showing better results than the task-specific models. Fig. 1 showcase the efficiency of our proposed LiteMuL with respect to the existing state-of-the-art solutions.

There exists literature on learning joint representations for both NER and POS using MTL and they have shown promising results in avoiding over-fitting but have been applied for server-side processing [8], [9]. Further, works on building on-device inference models for text classification [10], [11] highlight the advantages of on-device inferencing of neural models. Inspired by this work and due to lack of lightweight on-device models, we are proposing a novel on-device MTL model that automatically discovers necessary features for each task and work for the short informal conversational texts as well.

In our proposed LiteMuL model for NER and POS subtasks, we combine word-level and character-level representations to learn better representations of informal text. We use a shared BiLSTM layer to learn the common features and a NER-specific layer to learn task-specific features. The LiteMuL model is validated on our generated synthetic conversational text and the standard CoNLL 2003 dataset [12]. Our experiments show that NER and POS are benefitted from joint learning, with an improvement in accuracy and reduction in model-size, over the task-specific on-device models. This is mainly due to our

> [Fischler]PER proposed EU-wide measures after reports from [Britain]LOC and [France]LOC that under laboratory conditions sheep could contract Bovine Spongiform Encephalopathy (BSE)– mad cow disease.

Figure 2. Formal sentence sample

> It's a partyyyy
> This time at [Johnson's]PER new house
> Run, run, run!

Figure 3. Informal sentence sample

efficient MTL architecture design with minimal model parameters.

Fig. 2 & 3 show an example of formal and informal sentences, respectively.

Following are the major contributions of this paper:

- First, we propose on-device task-specific models for NER and POS which outperform the respective existing on-device models.

- We propose an efficient and lightweight on-device MTL neural network model (LiteMuL) for NER and POS with three variations: a) LiteMuL with Long short-term memory (LSTM) based char encoding (named as LiteMuL-LSTM), b) LiteMuL with Convolutional neural network (CNN) based char encoding (named as LiteMuL-CNN), and a) LiteMuL with Conditional Random Fields (CRFs) layer (named as LiteMuL-LSTM-CRF).

- The proposed models are benchmarked against publicly available CoNLL 2003 dataset. To benchmark our proposed models on informal conversations data, we also curated user data and extrapolated it with synthetic conversational sentences (see subsection IV-A). We also evaluate our proposed models on the system-specific metrics such as inference time and model-size (Refer subsection IV-B).

The experiments on LiteMuL-LSTM shows up to 11% accuracy improvement, 50%-56% model-size reduction, and 2-5% reduction in inference times in comparison to the independent models (See subsection IV-B). LiteMuL-CNN resulted in 41% reduction in inferencing time while maintaining comparable accuracy and model-size with respect to LiteMuL-LSTM. LiteMuL-CNN-CRF reduces the model-size and inference time further by 27%-30% and 5-10%, respectively with a slight improvement in accuracy in comparison to the LiteMuL-CNN (refer Table VIII).

The rest of the paper is organized as follows. In Section II, related work has been presented. Section III and Section IV give the proposed methodology and experimental analysis, respectively. Finally, Section V concludes the work and mentions the future scope.

## II. RELATED WORK

Sequence labeling tasks such as named entity recognition (NER), chunking, part-of-speech (POS) tagging, etc., are well-researched topics, many of which advocate the use of handcrafted features [13] derived from supervised machine learning approaches like Conditional Random Fields (CRF) [14]–[17], Hidden Markov Models (HMMs) [18], [19], Support Vector Machines (SVM) [20], [21], etc. These approaches usually do not adapt well to a new domain or task and often require a different set of handcrafted features even if two tasks are closely related like those of NER and POS tagging. Recent advancements in neural network architectures have outperformed traditional approaches, but most of these solutions proposed for linguistically related tasks are trained independently and infers information independently.

In literature, Sateli et al. [22] proposed a method to integrate NLP into smartphone applications by leveraging a web-service based Android library, where the actual NLP processing takes place on a server. However, such pipelines not only require a good network bandwidth but also come with the risk of exposing user data to third parties. To this end, researchers have explored the concept of on-device AI that can infer, and in recent works, even train [23], models exclusively in an on-device environment. Yet these approaches to NLP deal with specialized modules to perform specific tasks. In an end-user device, applications often need to use multiple such models simultaneously to achieve an end-goal, requiring each of them to take up more units of memory and processing power separately.

MTL has been shown to have improvements for several NLP tasks such as text classification [24], question answering systems [25], document ranking & query suggestion [26], etc. [27]. These works have proved that successful MTL applications help avoid local minima traps. Collobert and Weston [28] use MTL in a unified model to train multiple core NLP tasks: NER, POS tagging, chunking, and semantic role labeling with deep neural networks, showing that MTL improves generality among shared tasks. Furthermore, a good amount of research effort has been put forward towards improving the performance of NER systems using MTL approaches [29], [30]. In sequence labeling tasks, this can be seen as a form of inductive transfer that introduces an auxiliary task as an inductive bias to help a model prefer some hypotheses to others [31], [32]. This motivated us to explore on-device MTL for NER and POS tasks.

Many proposals in literature have focused on MTL for NER [33] or, POS [34], [35] or, both NER & POS [8], [9], [28], [36]–[39], but these solutions require huge RAM/ROM for on-device inferencing. Existing efforts towards employing MTL for tasks such as sequence labeling and semantic tasks have primarily focused on accuracy, leading to models that are huge for on-device use where resources are constrained [6], [40]. General-purpose models that emphasize on model-size still consume significant ROM: 200 MB (for DistilBERT [7]) and 119 MB (for MobileBERT [6] quantized int8 saved model and variables; sequence length 384). Transformer compression techniques like MiniLM [41] (12-layer, 33M parameters) also show a significant size of 68 MB with only 30k vocabulary. This poses a need to leverage MTL for NER & POS for on-device porting that outperforms strong baselines in terms of model-size and latency on smartphones.

## III. PROPOSED METHOD

In this section, we present LiteMuL – multi-task based on-device sequence tagger for NER and POS with reduced model-sizes. We first introduce the task-specific on-device models for NER and POS (see Fig. 5) to extract information separately. And then we describe the LiteMuL model (see Fig. 6) with multiple variants (LSTM/CNN for character representation and CRF over Softmax layer). The motivation behind trying three variants of LiteMuL is to experimentally achieve the best possible model when considering the device centric metrics (model-size and inference time). Before describing the model, we will discuss about data representation techniques that are used by our proposed on-device models.

### A. Data representation Techniques

A combination of word-level and character-level input representations has shown great success for several NLP tasks [42]. This is because word representation is suitable for relation classification but it does not perform well on short, informal, conversational texts whereas char representation handles such informal texts very well but does not perform well for long sentences. To take the best of both the representations, our proposed on-device models employ a combination of word and char representations for robust representations in case of formal & informal texts.

*1) Character level features:* Prior work has shown that incorporating contextual character-level representations of words can boost the accuracy of neural network models by handling both rare and misspelled words as well as model sub-word structures such as prefixes and endings. Major neural character language models include the character level CNN [43] and (Bi)RNN [44]. Fig. 4 depicts the char-level data representation techniques utilized by our proposed models. For character-level encoding, we employ two techniques: a) char LSTM and b) char CNN which encodes the input character, $c_i$ into $e_{c_i}$. We observe that our proposed MTL model with char CNN outperforms that of char LSTM (refer Table VII).

*2) Word Level Representation:* We use a similar setup for our context-sensitive word encodings as the character encodings. In this, we dynamically learn word embedding ($e_{w_i}$) for each word ($w_i$) in the corpus. Then, the CNN-based or LSTM-based character-level representation vector is concatenated with the word embedding vector to generate final word representations ($o_w$):

$$o_{w_i} = \text{concat}[e_{w_i}, F(e_{c_1}, e_{c_2} ..., e_{c_n})]. \quad (1)$$

Where, F is LSTM or, CNN.

### B. Independent Model Architecture for NER and POS Tasks

The independent model architectures followed by our independent on-device NER and POS models are shown in Fig. 5. For char representations, we use Char LSTM which is concatenated with the word representation. The concatenated embeddings are fed into another level of word-level Bi-LSTM layer to process the entire text sequence as illustrated in Fig. 5. Finally, the output vectors of the Bi-LSTM layer are fed to a fully connected layer with the Softmax activation function which outputs the probability of each defined label for each input.

*1) On-device Independent NER Model Specifications:* After hyperparameters tuning, we fix the maximum sequence length to 30 and maximum character length to 15. We set character and word embedding size to 6 & 12 for CoNLL data and 8 & 16 for custom data, respectively. Thereafter, a Time Distributed Character LSTM layer with 10 units (i.e. the dimensionality of the output space) generates character encoding which is concatenated with word embedding. Also, a Spatial Dropout layer of rate 0.3 is utilized to reduce overfitting. Finally, we added a word-level Bi-LSTM layer with 20 units having a recurrent dropout rate of 0.6. The final dense layer with Softmax activation outputs a probability distribution over NER labels. The network training is done with a batch-size of 64 and compiled with Adam optimizer [45] using a sparse categorical cross-entropy loss function. The model converges after 95 and 10 epochs for CoNLL and custom datasets, respectively. We have chosen hyperparameters to optimize the accuracy while maintaining the small model-size.

From Fig. 6, we can observe an increasing trend in accuracy and model-size with respect to embedding size.

*2) On-device Independent POS Model Specifications:* Along with the CoNLL test set, we also utilize treebank from Universal Dependencies (UD) v2.2 [46] for model evaluation. The data labeling is updated based on Penn Treebank (PTB) tags, in which we support 36 tags (reduced from 45 to 36 tags after merging different punctuations and symbols into a single tag). The total number of trainable parameters are only ~204,000 for uncased model trained on CoNLL data. After hyperparameters tuning, we set the character and word embedding dimensions to 6 and 8, respectively. The character embeddings are passed to an LSTM layer with an 8-dimensional output vector which yields the contextualized character-level word encodings. After generating the concatenated embeddings, we add a Spatial Dropout layer with a rate of 0.1. In the case of word-level BiLSTM, we are using regular and recurrent dropouts, each with a rate of 0.2. The training of the whole neural network is conducted using the default batch-size of 32, and 17 epochs. Additionally, we compile the model using the Adam optimizer [45]. We observe a similar plot for the effect of embedding dimension on accuracy and model-size (as described in Fig. 6 for the NER model).

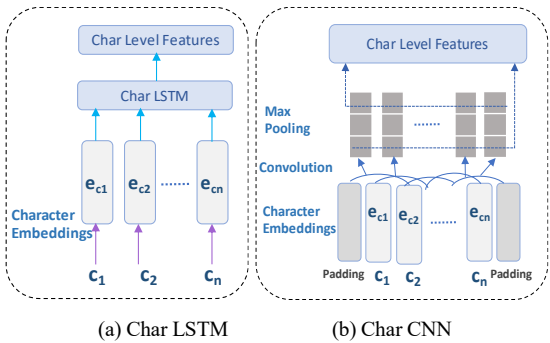

Figure 4. Illustration of char representation techniques.

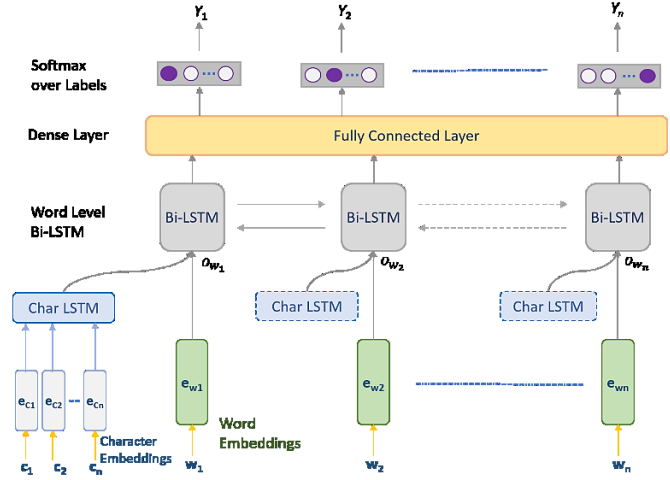

Figure 5.  Illustration of our Proposed Model Architecture for POS and NER Predictions.

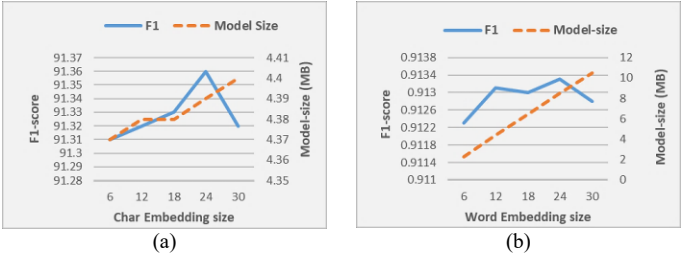

Figure 6.  The effect of embedding dimension on NER accuracy and model-size for CoNLL data (Cased).

*C. LiteMuL*

Our proposed LiteMuL model architecture, developed to learn NER task and POS tagging task together, is depicted in Fig. 7. In this configuration, for the given input sequence $W= [w_1, w_2, .., w_n]$ containing $n$ tokens, the main task is NER ($Y^{NER}= Y^{NER}_1, Y^{NER}_2, …, Y^{NER}_n$) and the auxiliary task is POS tagging ($Y^{POS}= Y^{POS}_1, Y^{POS}_2, …, Y^{POS}_n$). Both subtasks share a common data representation layer described in Section II to extract the useful features. The output of the data representation layer is fed into a shared BiLSTM layer which serves as input for the task-specific (BiLSTM) layer of NER and a fully connected layer of POS. For each task, an independent Softmax layer is employed at the output of the respective fully connected layers to predict the class labels with the highest probability for both the subtasks.

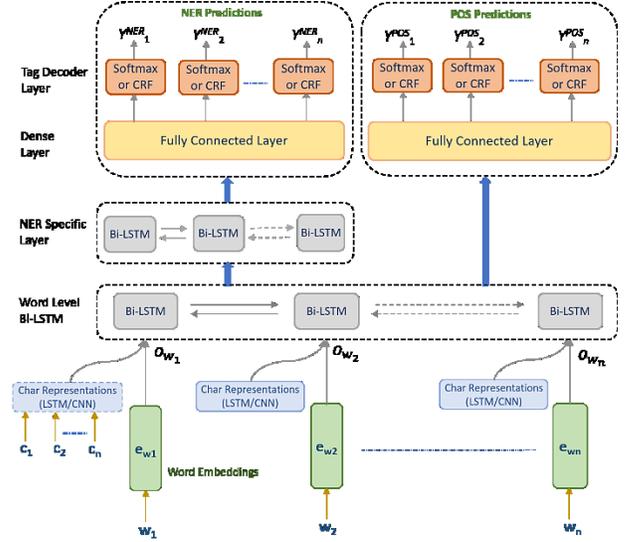

Figure 7.  The Proposed LiteMuL Model Architecture for POS and NER Predictions.

In literature, Conditional Random Fields (CRFs) has shown good accuracy for sequence labeling tasks. It uses the weights of the dense layer to perform a sequential classification. Therefore, we exploit CRF on top of the Dense layer. The comparative result analysis of Softmax and CRF layers presented in experimental section demonstrates superiority of CRF over Softmax for NER and POS.

In this configuration, both the subtasks are jointly learned on the same dataset using a joint training loss-function, defined below:

$$J_{LOSS}= W_{NER} L_{NER} + W_{POS} L_{POS}. \quad (2)$$

We take a weighted sum of the losses from both NER and POS subtasks by assigning 1 and 1.5 weights, respectively. Here, $L_{NER}$ and $L_{POS}$ are loss functions (categorical cross-entropy) for NER and POS subtasks, respectively and $W_{NER}$ and $W_{POS}$ are weights associated with them. These weights are selected by our empirical analysis performed on the CoNLL train and development sets.

In our experiment, we train the model with 64 batch-size. We find the optimal dimensions for the character and word embedding to be 6 & 12 (CoNLL data) and 8 & 16 (custom data), respectively. We set the char-level LSTM size to 10 with regular and recurrent dropouts, each with a rate of 0.2 and 0.5, respectively. Similarly, we set the size of the word-level BiLSTM to 20, each with a recurrent dropouts of 0.6. The model converges after 95 epochs and 10 epochs for CoNLL and custom datasets, respectively. The total number of trainable parameters vary in the range of 130,979 to 312,937 for all the MTL model results reported in this paper.

We train our proposed models on the train-set of the CoNLL 2003 dataset and use CoNLL train-development sets for the hyperparameters tuning.

## IV. EXPERIMENTAL RESULTS

In this section, we present our proposed techniques for dataset generation and then showcase the result analysis.

### A. Dataset

Current approaches on entity detection on resource-constrained devices such as mobile phones are limited to detect telephone numbers, URLs using regular expressions. The publicly available entity detection datasets such as CoNLL2003 is a collection of news-wire articles. Although the ACE2005 dataset contains samples on life events, the categories defined and the type of text are not suitable for our problem domain.

For comparison with respect to existing works, we benchmarked the proposed model against the standard datasets, CoNLL 2003 [12] & Universal Dependencies [46]. Besides, for measuring the model's effectiveness, in short, informal text, we evaluate the model on our custom-generated dataset.

*1) Public Dataset:* We are using CoNLL 2003 data, which is most commonly used for sequence labeling tasks. It consists of standard train, development, and test sets which supports 8 NER tags: I-MISC, I-ORG, B-MISC, I-PER, B-LOC, B-PER, B-ORG, I-LOC, and O. For testing POS prediction accuracy, we also use the Universal Dependencies dataset which is widely used in literature for POS benchmarking.

*2) Custom-generated Dataset:* As mentioned in the introduction, the main motivation for the proposed idea is to perform well for both formal and informal conversations. The majority of the off-shelf NER systems were trained on news articles & Wikipedia genre text and they often fail in extracting the entities from the short-hand text of users. In-order for the model to be robust for the shorthanded, intelligible, inconsistent spellings, we have curated the corpus from a user trial of 100 users spanned over 2 weeks. The curated conversations consist of multiple categories of data like regular chitchats, daily wishes, sharing quotes, and sending impromptu invites.

Since the neural networks are data-hungry and the collected datasets are not that sufficient in quantity, we have extrapolated the data synthetically. We have chosen to focus on the following categories (see Fig. 8) to enable other key applications in smartphones (like auto-creating calendar reminders, sharing location markers, etc.).

The user conversations are curated along with annotations. To be in-line with CoNLL dataset annotations, the user annotations are converted to IOB format programmatically.

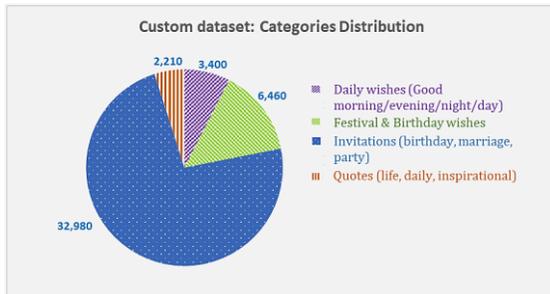

Figure 8. Custom-generated Dataset Statistics

### B. Result Analysis

In this section, we report micro-averaged F1-score (referred to as F1) for NER model whereas accuracy-score for POS model. We report the accuracy results for the proposed models trained in two different settings (a) cased and (b) uncased for both CoNLL 2003 and custom datasets. We refer to the custom-generated train-set as custom-train, US test-set as US-test, and Indian test-set as Ind-test. In the uncased, all the texts are getting converted to the lower-case before feeding to the model. We use the Tensorflow framework [47] for building the models.

As the proposed solution is mainly for mobile devices, we have also evaluated system specific metrics (model inference time in milliseconds & model-size in MB) which are reported in this Section. By inference time, we imply the average time taken by the neural model to process an input word sequence and return the labeled output. Experiments for model inferencing are conducted on the Samsung Galaxy Note8 device. Initially, we set the maximum sequence length that the model can process at a time to 75. However, we observe that the average length of input sentences in our data is ~30, and processing a sequence length of 75 for smaller sentences is computationally expensive. To overcome this, we set the sequence length to 30. After hyperparameters tuning, we observed that some of the parameters vary across NER, POS, and LiteMuL models for CoNLL 2003 and custom datasets.

First, we benchmark our independent models against the respective state-of-the-art solutions (refer to Table I & II). We compare the results of the independent NER model with the existing state-of-the-art NER models: OpenNLP and SpaCy. OpenNLP provides a separate model for each entity, whereas we encapsulate that functionality into a single model. OpenNLP model doesn't differentiate between B and I tags, thus for comparison we combine B and I tags and compare with the corresponding results of OpenNLP NER. We also evaluate the on-device inference time of the independent NER model with respect to the OpenNLP NER (see Table I). In Table II, we benchmark our independent POS model with the current state-of-the-art POS models, OpenNLP, and SpaCy. We observe that though our POS models outperform OpenNLP in model-size and accuracy.

Table III compares the NER predictions of our LiteMuL-LSTM (NER_MTL) with the independent NER model (NER_IND) for both the datasets in different settings. Similarly, we also compare the POS predictions of our LiteMuL-LSTM (POS_MTL) with respect to the independent POS model (POS_IND) for both the datasets (see Table IV). Table IV shows that POS_MTL accuracy is better than POS_IND in most of the cases (except custom uncased). We can infer from Table III & IV that NER and POS accuracy obtained by LiteMuL-LSTM with cased text (without lower-casing) gives higher accuracy as compared to the uncased version.

Table V demonstrates the model-size of independent models and LiteMuL-LSTM model for both CoNLL and custom datasets with cased and uncased versions. It is evident from the results that the LiteMuL-LSTM reduces the overall model-size by 32-38%.

Further, we can see that our LiteMuL-LSTM model inference time is less than the total inference time of NER and POS subtasks combined (refer Table VI). CoNLL data contains both NER and POS tags that is the reason of considering it for this analysis.

Finally, we report the accuracy results and relevant metrics for CNN-based encoding in Table VII. Here also, we observed that the best accuracy is achieved without lower-casing (cased). We can see that CNN-based encoding results in 41% reduction in inferencing time in comparison to LSTM-based encoding (refer Table VI & VII) with 1% increase in memory footprint (refer Table V & VII). The model-size reduction can be attributed to the fact that CNNs leads to a reduction in complexity by parallelly concentrating on the key features.

We also report the accuracy results and relevant metrics for CRF-layer with CNN-based encoding in Table VIII. Model-size has been reduced by 27% and 30% for custom data and CoNLL data, respectively after applying the CRF layer whereas accuracy is improved slightly except for POS CoNLL accuracy trained over CoNLL-train cased dataset.

TABLE I. COMPARISON OF OUR INDEPENDENT NER MODEL WITH OPENNLP AND SPACY

| Model | Model-size (in MB) | F1-score | | | Inference Time (in ms) | | |
|---|---|---|---|---|---|---|---|
| | | CoNLL | US-test | Ind-test | CoNLL | US-test | Ind-test |
| NER | 3.51(CoNLL)/ 1.56(Custom) | **0.92** | **0.98** | **0.96** | 15 | 21 | 22 |
| OpenNLP | 24.1 | 0.75 | 0.68 | 0.67 | 23 | 66 | 67 |
| SpaCy | 13 | 0.89 | 0.93 | 0.92 | Offline | | |

TABLE II. POS MODEL BENCHMARKING WITH RESPECT TO OPENNLP AND SPACY

| Model | Model-size (in MB) | Test-set Accuracy | | Inference Time (in ms) | |
|---|---|---|---|---|---|
| | | CoNLL | UD | CoNLL | UD |
| POS_1 | **2.35** | 0.8746 | 0.8263 | 21 | 20 |
| POS_2 | 5.10 | **0.9030** | **0.8980** | 21.5 | 21 |
| OpenNLP | 5.50 | 0.9021 | 0.8912 | **4.5** | **4.5** |
| SpaCy | 11.00 | 0.7865 | 0.8620 | Offline | |

TABLE III. NER PREDICTIONS ACCURACY COMPARISON OBTAINED BY INDEPENDENT NER MODEL AND LITEMUL-LSTM MODEL

| Train-set | Casing | Test-set | NER_IND F1 | NER_MTL F1 |
|---|---|---|---|---|
| CoNLL-train | Uncased | CoNLL-test | 0.9197 | **0.9245** |
| | Cased | CoNLL-test | **0.9490** | 0.9451 |
| Custom-train | Uncased | US-test | 0.9804 | **0.9879** |
| | | Ind-test | 0.9583 | **0.9764** |
| | Cased | US-test | 0.9792 | **0.9871** |
| | | Ind-test | 0.9553 | **0.9714** |

TABLE IV. ACCURACY RESULTS COMPARISON FOR POS PREDICTIONS USING INDEPENDENT MODEL AND LITEMUL-LSTM MODEL

| Train-set | Casing | Test-set | POS_IND Accuracy | POS_MTL Accuracy |
|---|---|---|---|---|
| CoNLL-train | Uncased | CoNLL-test | 0.8746 | **0.8989** |
| | | UD-test | 0.8263 | **0.8331** |
| | Cased | CoNLL-test | 0.9075 | **0.9356** |
| | | UD-test | 0.8685 | **0.8726** |
| Custom-train | Uncased | UD-test | **0.6699** | 0.6565 |
| | Cased | UD-test | 0.6704 | **0.7099** |

TABLE V. MODEL-SIZE (IN MB) COMPARISON ANALYSIS FOR NER MODEL (NER_IND), POS MODEL (POS_IND), AND LITEMUL-LSTM MODELS

| Train-set | Casing | Model-size (in MB) | | | |
|---|---|---|---|---|---|
| | | NER_IND | POS_IND | NER_IND+ POS_IND | MTL |
| CoNLL-trains | Uncased | 3.15 | 2.35 | 5.50 | **3.41** |
| | Cased | 3.51 | 2.59 | 6.10 | **3.78** |
| Custom-train | Uncased | 1.51 | 1.04 | 2.55 | **1.73** |
| | Cased | 1.56 | 1.07 | 2.63 | **1.77** |

TABLE VI. INFERENCE TIME (IN MILLISECONDS) COMPARISON ANALYSIS OF NER_IND, POS_IND, AND LITEMUL-LSTM MODELS

| Train-set | Casing | Inference Time (in ms) | | | |
|---|---|---|---|---|---|
| | | NER_IND | POS_IND | NER_IND+ POS_IND | MTL (NER+POS) |
| CoNLL-train | Uncased | 15 | 20 | 35 | **34** |
| | Cased | 15 | 21 | 36 | **34** |

TABLE VII. LITEMUL-CNN PREDICTIONS, INFERENCE TIME (IN MILLISECONDS), AND MODEL-SIZE (IN MB) ANALYSIS

| Train-set | Casing | Test-set | NER F1 | POS Accuracy | Inference Time | Model-size |
|---|---|---|---|---|---|---|
| CoNLL-train | Uncased | CoNLL | 0.9387 | 0.9035 | 19 | 3.43 |
| | | UD-test | NA | 0.8420 | 20 | |
| | Cased | CoNLL | 0.9423 | 0.9409 | 19 | 3.80 |
| | | UD-test | NA | 0.8794 | 20 | |
| Custom-train | Uncased | UD-test | NA | 0.6347 | 22 | 1.75 |
| | | US-test | 0.9831 | NA | 21 | |
| | | Ind-test | 0.9709 | NA | 21 | |
| | Cased | UD-test | NA | 0.7431 | 22 | 1.79 |
| | | US-test | 0.9895 | NA | 21 | |
| | | Ind-test | 0.9778 | NA | 21 | |

TABLE VIII. LITEMUL-CNN-CRF PREDICTIONS, INFERENCE TIME (IN MILLISECONDS), AND MODEL-SIZE (IN MB) ANALYSIS

| Train-set | Casing | Test-set | NER F1 | POS Accuracy | Inference Time | Model-size |
|---|---|---|---|---|---|---|
| CoNLL-train | Uncased | CoNLL | 0.9433 | 0.9090 | 17 | 2.39 |
| | | UD-test | NA | 0.8575 | 17 | |
| | Cased | CoNLL | 0.9562 | 0.9391 | 17 | 2.63 |
| | | UD-test | NA | 0.8813 | 17 | |
| Custom-train | Uncased | UD-test | NA | 0.6732 | 20 | 1.27 |
| | | US-test | 0.9917 | NA | 20 | |
| | | Ind-test | 0.9843 | NA | 20 | |
| | Cased | UD-test | NA | 0.7461 | 20 | 1.30 |
| | | US-test | 0.9922 | NA | 20 | |
| | | Ind-test | 0.9833 | NA | 20 | |

*C. Discussion*

Our proposed LiteMuL model for sequence tagging outperforms the existing sequence tagging solution by reducing resource-centric metrics: model-size & inferencing time while maintaining accuracy. This is due to the shared model architecture and joint learning through multi-tasking. Our proposed model outperforms OpenNLP NER (on-device model) in terms of accuracy, model-size, and inference time. From the reported results we deduce that in most of the cases NER and POS accuracy obtained with cased text (without lower-casing) gives better accuracy as compared to the uncased version. LiteMuL-CNN reduced inference time by 41% in comparison to the LiteMuL-LSTM (refer Table VII). Our proposed LiteMuL-CNN-CRF model reduces the model-size and inference time further by 27%-30% and 5-10%, respectively with a slight improvement in accuracy over the LiteMuL-CNN model (refer Table VIII). We compared our three variants of LiteMuL and found that LiteMuL-CNN-CRF outperforms in terms of all three parameters: model-size, accuracy, and inference time. This enables LiteMuL-CNN-CRF for on-device deployment without memory restrictions.

V. CONCLUSION

In this paper, we developed lightweight, fast, and accurate LiteMuL model with three variations that can fit on mobile devices for the entity and part-of-speech tagging by capturing relevant features from short, informal user text. Our novel approach utilizes the shared data representation layer and BiLSTM layer, resulting in up to 3.80MB memory footprint with maximum inference time of 34 milliseconds and 22 milliseconds for LiteMuL-LSTM and LiteMuL-CNN, respectively. Our proposed LiteMuL-CNN-CRF model further reduces the model-size and inference time to 2.63MB and 20 milliseconds, respectively. Hence it is very efficient for on-device deployment with optimal memory usage. Experiments on standard benchmarking datasets (CoNLL 2003, Universal Dependencies) and custom-curated datasets showed that our model improved upon independent models in terms of size reduction, inference time, and accuracy metrics.

In the future, we want to extend this approach to process multi-lingual short, informal conversational text. Besides, we plan to further optimize our models via quantization and model-size reduction using Tensorflow Lite.